# Looking forward: Linguistic theory and methods

*John Mansfield and Ethan Gotlieb Wilcox*
University of Zurich and Georgetown University

**ABSTRACT**

This chapter examines current developments in linguistic theory and methods, focusing on the increasing integration of computational, cognitive, and evolutionary perspectives. We highlight four major themes shaping contemporary linguistics: (1) the explicit testing of hypotheses about symbolic representation, such as efficiency, locality, and conceptual semantic grounding; (2) the impact of artificial neural networks on theoretical debates and linguistic analysis; (3) the importance of intersubjectivity in linguistic theory; and (4) the growth of evolutionary linguistics. By connecting linguistics with computer science, psychology, neuroscience, and biology, we provide a forward-looking perspective on the changing landscape of linguistic research.

## 1. INTRODUCTION

William Labov's festschrift is titled *Towards a Social Science of Language* (Guy et al. 1996), while Noam Chomsky's book of interviews is *The Science of Language* (Chomsky 2012). Linguistics has long been preening itself for scientific status, and in this chapter we examine some ways the field continues to pursue a scientific understanding of humanity's most enigmatic gift. As we will show below, the use of computational methods and large datasets are currently driving advances in linguistics, providing more accurate (or at least reproducible) evidence on our major theoretical questions.

Much of the credit for progress lies with increasing connections to other disciplines. We here advocate for a linguistics that is richly connected with computer science, psychology, neuroscience and biology. In the account that follows, we will especially highlight connections to computational linguistics, since some of the most startling recent developments in computer science focus on language data. Inevitably, our account is very partial, and makes no claim to objectivity. Our own areas of research undoubtedly shape the themes selected for attention here, though we have attempted to cast a reasonably broad net over recent developments.

We discuss four main themes below, while also aiming to show some common threads running between them. Firstly, we observe a recent trend for explicit testing of basic hypotheses about linguistic representations. Encompassing such principles as efficiency, locality, and conceptual grounding, we outline a type of research that aims to evaluate evidence systematically against cross-linguistic datasets, in order to provide more scientifically grounded answers to questions about the basic nature of human grammars. Space constraints here limit our focus primarily to grammatical structure, though similar trends could also be reported for phonology, morphology and semantics.

Secondly, we discuss the recent rise of artificial neural network methods. These differ fundamentally from traditional linguistic theory because their computations involve sub-symbolic interactions between numerical functions, whereas language structure is instead usually represented as computation of discrete symbols. However, since neural networks have become capable of mirroring human linguistic production with alarming accuracy, they have inevitably become the focus of theoretical debates in linguistics, while also offering new methods for analysing data.

Thirdly, we turn to the question of intersubjectivity, that is, the fact that human language use occurs not in the context of disembodied symbols, but rather in the interaction of minds. Linguistic structure is to a great extent shaped by our mental states and our inferences about the mental states of others. We argue that the importance of this perspective is still growing in linguistics, since the subfield that deals with this problem explicitly—pragmatics—emerged much later than traditional studies of syntax, morphology and phonology.

Finally, we discuss the current renaissance of evolutionary linguistics, benefitting from intensified exchanges with biology and neuroscience. This approach to linguistics goes beyond static, discrete grammars, to create dynamic models of language change. One notable development here is an increase in phylogenetic modelling, which treats each language as a dynamic system, branching off from other languages in family trees. But some evolutionary research also models interpersonal variation among individual agents, as opposed to the more traditional abstraction of each 'language' as a homogeneous shared code.

Running through all four of these themes is the advance of computational methods and accessible data. However we also try to highlight the ways in which they are connected by their key questions and theoretical concerns. Testing hypotheses about symbolic representations is now aided by neural network, also called *connectionist*, models; one of the main critiques about neural networks is that they lack grounding in intentional agents, though this is also an active area of computational research. The study of linguistic evolution can itself be framed as the study of how intentional agents learnt to align their subjective states, and we will see below that some work now models such processes in a preliminary way.

Inevitably, there are many exciting areas of research that cannot included in this brief tour. Neurolinguistics, psycholinguistics, sociolinguistics and other subfields could contribute much to this discussion, but thankfully are covered in other sections of this book (see chapters XX, XX, XX). What we offer below is a selection of research that can be taken together as a coherent snapshot, or so we hope. While this is just a snapshot, we hope that it provides a sense of perspective for linguists who have watched the development of the discipline unfold, and an exciting entree for those who are new to the field.

## 2. THEORIES OF SYMBOLIC REPRESENTATION

Linguistic theory examines the nature of symbolic representations, with particular attention to how these occur in human languages and, by extension, the human mind. Linguistics focuses on human language in particular, as opposed to artificial languages, computer languages, etc. But when it comes to linguistic *theory*, it can be beneficial to consider human languages in the light of symbolic representations more generally, so that we may understand not just what human languages are like, but also how they fit into the broader possibility space. For example, we may find that there are mathematically optimal forms of representations which, however, are demonstrably distinct from human languages. Current linguistic theory continues to investigate general principles of symbolic representation, such as efficiency, locality, and transparency.

## 2.1. Rebalancing theory and observation

While there has been substantial continuity in theoretical principles of symbolic structure, there is also a rebalancing of the roles of theoretical assumptions and empirical observations. For some, this has been accompanied by a change in disciplinary nomenclature, with 'language sciences' being preferred as a way of emphasising the rigour of empirical methods. For others this has meant a deeper integration of empirical methods into all aspects of linguistics, not just in experiments that can substantiate theoretical claims, but in the judgement data used to outline and initially support theoretical claims themselves (e.g. Mahowald et al. 2016). The 'language sciences' movement amounts to more than a superficial rebranding: for example, thirty years ago, Abney (1996) published a programmatic article arguing that statistical methods, probabilistic concepts and gradient methods should be given more scope in linguistics. At the time, he observed that these were popular in computational and corpus linguistics, but still treated as largely alien to the core of 'theoretical' linguistics. To a large extent, Abney's call has been heeded, with a high proportion of studies in major linguistics journals now using some variety of statistical methods, often accompanied by code and data. There are also more studies of language appearing in high-profile, general-science journals.

Although the increase in empirical rigour is clear, it should be seen as a change in emphasis, rather than a sharp break. Empirical work is always entwined with *a priori* theoretical considerations, selection of formalisms, and delineation of the scope of enquiry. Empirical results are always in a conversation with theoretical foundations.

In the following sections we will review some of the general principles that play major roles in current linguistic theory. None of the principles we discuss are new, and all draw on longstanding traditions. However, if a trend is to be observed, it is that theoretical questions and toolkits are increasingly drawn from, and in conversation with, allied disciplines such as statistics, computer science, and cognitive sciences.

## 2.2. Efficiency and minimality

The concept of efficiency connects to many threads of contemporary linguistic research, and is increasingly used in formal applications drawing on information theory (Gibson et al. 2019; Levshina 2022). Efficiency is a broad concept, which can include any form of cost/benefit tradeoff, whereby the formal complexity of a representational system is minimised with respect to the utility of the message (i.e., semantic content) expressed. Often we think about efficiency in terms of length of words, number of morphemes, or some other enumeration of symbolic units. But efficiency can also apply to other dimensions of a representational system. Fewer rules, or fewer applications of rules, is another way of reducing complexity (Goldsmith 2001; Culicover & Jackendoff 2005). Therefore efficiency is not just about 'surface forms', but also connects to considerations of parsimony and simplicity in our models. Seen in this light, the theme of efficiency is akin to the 'minimalist' spirit that seeks to reduce the number of rules required to model language data (Chomsky & Lasnik 1993). Below we will return to the question of evaluating grammars, while in this section, we discuss one foundational issue in the question of efficiency – word lengths – and describe how the theoretical understanding of this issue has evolved, particularly in response to more sophisticated use of statistics in theory building, and better corpus resources for theory testing.

The foundational study of efficient representation measured the length of words against their frequency, showing that more frequent words are shorter, which overall reduces the cost or effort of communication (Zipf 1935). Further developments reconceptualised frequencies as *probabilities* of words appearing, with an inverse relationship between probability and length, and probability taken as the basis for mathematical formulations of information quantity (Shannon 1948). The intuition is that highly frequent expressions do not carry much new information, since they are already expected. Therefore less probable messages are more informative, and the greater length of such messages corresponds to a greater quantity of information.

The amount of information conveyed by an expression is not merely a function of its overall probability, but rather its probability in context. A word like *Francisco*, which is extremely rare and therefore highly informative in English, is highly predictable and much less informative given the context "I left my heart in San…" This notion of in-context information of words and expressions is captured by the information-theoretic metric *surprisal*. Surprisal is the negative log probability of a word, given its preceding context; a word that has a probability of zero has an infinite amount of surprisal (i.e., it conveys the maximal possible amount of information), and a word that is obligatory has a surprisal of zero (i.e., it conveys no information; it is entirely predictable). Turning back to our discussion of efficiency, Piantidosi and colleagues (2011), argue that rather than being efficiently packaged with respect to raw frequency, word lengths should be packaged with respect to surprisal, given that words occur in contexts. They find that surprisal is more predictive of word lengths than raw frequency counts across a variety of typologically distinct languages.

The extent and sophistication of research on linguistic efficiency has exploded in recent years, thanks to the boom in computational resources and techniques. For example, while twenty years ago, corpora that were considered large had word counts in the tens of millions (Barth & Schnell 2022), some corpora today have tens of billions of words. In addition to corpus size, Natural Language Processing techniques have produced more powerful statistical estimators. The most prominent current example are (large) language models (LLMs), which can provide accurate estimates of a word's probability, given its context. The dual advance of larger corpora and better statistical modeling techniques have allowed for much more accurate estimation of long-tail events, such as probabilities of rare words or constructions. This has again brought advances in the study of word lengths. While Piantidosi and colleagues (2011) improve over Zipf's (1935) efficiency hypothesis by factoring in context, Pimentel and colleagues (2023) propose a further refinement. They argue that word lengths are not only sensitive to context but also to the variance of information between contexts. In addition, they test their hypothesis by obtaining estimates of surprisal from large language models, trained on large corpora, improving over the count-based methods used in previous studies. These new methods and corpora give a surprising result—-word lengths are best explained by raw frequency after all, meaning that Zipf's initial observation stands strong.

We can also model efficiency in ways that go beyond the predictability of words in context. For example we can ask, how well can an addressee decode the compositional semantics of a sentence, given the form of that sentence? One recent study, taking advantage of recent multilingual corpus collections and parsing technologies (UD Pipe 2017), focuses on the semantic roles of referents as agents, patients, etc, testing how much word order and case marking contribute to distinguishing these roles (Levshina 2021). We can consider word order and case marking to be 'informative' inasmuch as they are consistently associated with particular semantic roles, and this perspective forms the basis for evaluating how much these might help a hypothetical addressee to decode sentences. Findings show a trade-off between case and word order, where one or the other tends to provide a more robust cue for semantic roles. Additional causal analysis suggests that presence of rigid word order tends to precipitate loss of case markers. When we consider these findings in relation to the 'complexity of rules' issue mentioned above, they suggest efficiency in minimising the amount of structure being used to express semantic roles.

Levshina's results suggest that each language makes a complexity tradeoff. While a certain amount of complexity is necessary to successfully communicate a complex meaning, principles of efficiency will keep languages from becoming overly complex, by trading off one area of complexity for another. However other studies find inefficiencies, and cross-linguistic differences of efficiency. Mahowald and colleagues (2023) find substantial redundancy in the cues identifying subjects of clauses. Recently, another study (Koplenig et al. 2023) used LLMs to estimate the overall information content of different languages, as a test of the complexity tradeoff hypothesis. They find that languages are not equally complex, and that greater complexity correlates with larger speaker populations. However, their findings draw to a large extent on translated texts, which could introduce a confound if the process of translation tends to reduce complexity.

## 2.3. Implementing and evaluating models of grammar

A long-standing challenge of linguistic theory is how to evaluate any formal representational system against natural linguistic data (Chomsky 1957: 51). Over the last couple of decades, as it becomes more feasible to computationally implement grammars, this challenge has become tractable. It is now possible to implement a specific theoretical proposal as a computational model that makes predictions about linguistic data, and to evaluate these predictions against a large corpus. The information-theoretic measures mentioned above often play an important role in the evaluation of a model's success. But as we will see below, there is also some recent work where information-theoretic concepts become part of the model itself, that is to say, reformulating models of syntax in information-theoretic terms.

Predictive computational modeling usually focuses on challenges such as how to output a set of words in the right order, or how to assign the right hierarchical structure to a sequence of words. Models often 'learn' such principles from input corpus data, and can then be evaluated on how well they can replicate corpus data after such training. There is generally some type of statistical learning mechanism, in some cases involving an artificial neural network. In these networks the symbolic inputs and outputs are mediated by substantial non-symbolic machinery, a question to which we return below. For example, earlier models of language production showed that a moderate level of output accuracy could be obtained simply from probabilistic word-word sequences in a smallish input corpus (Chang et al. 2005), or achieving greater accuracy by also incorporating labels for thematic roles (Chang et al. 2006). These studies highlighted the potential for explicitly evaluating the output of grammars with different levels of complexity.

Earlier work tended to focus on modelling a single language, usually English, but more recent work has evaluated the potential for a single model to achieve accurate representation for a range of diverse languages. One recent study (Hahn et al. 2020) develops an algorithm for ordering word-classes such as noun, verb and adjective, to test whether such orderings match word-order correlations previously proposed by Greenberg (1963). To test Greenberg's proposals against multi-lingual corpus data, they measure the 'informativity' of the grammar, which is higher to the extent that this linearisation helps decode the underlying dependency structures of sentences (i.e. makes dependency structure more predictable, given a surface word order). This allows them to discover the most informative grammars, based on the word-ordering model, with the striking result that the optimal grammars match all of the Greenbergian grammar principles they tested. This is a good example of how high-level theoretical claims can now be tested, using statistical methods that explicitly model principles of communication.

## 2.4. Locality and harmony

The general idea of locality is that symbols that must be composed, such as a noun and its modifying adjective, or a verb and its argument, should be close together in some way (Hawkins 1994; Hawkins 2004; Gibson 1998). This is often operationalized in terms of linear proximity, with symbolic composition represented as dependency grammar annotations. But phrase-structure grammars have similar principles in demanding that certain types of symbolic composition must occur 'locally' within certain subtrees (Rizzi 2013). Like the efficiency principles discussed above, locality can be seen as a principal of optimisation for grammar, here applying to hierarchical compositionality, rather than to brute quantities of expressive material. Non-local composition is theoretically more 'costly', for a human mind, a computer program, or in the evaluation of a formal grammar. For example, a parser or compiler may need to store symbols in some kind of memory buffer as part of the production or parsing process, and we might assume that there is a cost associated with the number of such symbols.

Psycholinguistic experiments have found support for locality preferences in both language production and comprehension (Bartek et al. 2011) . But the recent development of large corpora for multiple languages has revealed new types of evidence. For example, dependency-tagged corpora can be used to calculate the degree of locality in natural languages, enabling a test of the hypothesis that locality is driving syntactic linearisation. One study compares natural corpora against artificially permuted counter-factual corpora, showing that the natural corpora consistently have much better locality scores than permuted alternatives (Futrell & Levy & et al. 2020). A further study develops more rigorous counter-factual comparisons, by taking into account the head-initial or head-final tendencies of natural languages, and constraining the permuted corpora to match the natural data in this respect (Jing et al. 2022). With this more conservative test in place, they still find evidence for a locality preference in some languages, but also find exceptions, in particular some languages with a strong tendency towards head-final syntactic ordering.

Locality constraints take a more absolute form in the theory of island constraints on filler-gap dependencies. Filler–gap dependencies (in English) are the relationship between a wh- element, such as *who* or *what*, and a gap, which is an empty syntactic position. Here, the dependency is potentially unbounded, but constrained by types of hierarchical structure. For example, filler–gap dependencies can span multiple clauses, as in (1a), but are not allowed when the gap occurs inside a relative clause, as in the ungrammatical (1b), below:

(1)     a. *I know who the journalist said their source spoke to __ last week.*
        b. * *I know who [the journalist that disliked __ ] asked the rude question.*

Islands are cases of locality effects; the filler and the gap can co-occur only if they appear in some 'local' domain. But what exactly that domain is, and how it is learned is a matter of ongoing debate. Traditionally, island effects like the one in (1) are treated as being disallowed by the grammar, with the acceptable domains for the effects being parameterized early during language acquisition. However, alternative theories posit that the island effects are not grammatical, but rather byproducts of processing difficulty. An alternative approach posits that island effects result not from purely syntactic constituent structure, but from infelicitous use of foregrounding (Ambridge & Goldberg 2008). Wh- elements near initial position, such as *who* in (1a), are used to foreground a referent as being 'at issue' in the current exchange, but backgrounded position such as the relativised object in (1b) are used for presupposed referents. The infelicity of (1b) can therefore be explained by a clash of foregrounding and backgrounded, as supported in recent experimental studies (Cuneo & Goldberg 2023).

Recently, several studies have asked whether islands can fall out of efficient processing. Wilcox et al. (2024) demonstrate that many of the island effects in English are learned by LLMs, even though the models (presumably) have no experience with islands in their training data. They take these results to demonstrate that abstract concepts such as local domains and syntactic dependencies can be learned as a byproduct of predictive processing.

Other studies of linear locality have proposed an alternative formulation, where the affected word pairs are not those that have a syntactic dependency relation, but instead an informational relationship, namely a high degree of Mutual Information. This is a measure of how strong the 'association' is between two words, but more specifically, it measures how information each word provides about the other. This version of locality theory is therefore labelled 'information locality', and again finds empirical support in corpus studies (Futrell 2019; Culbertson & Schouwstra & et al. 2020) (Futrell 2019; Culbertson et al 2020). Dependency locality theory assumes that syntactic dependencies should be composed with a minimum of intervening material, while information locality assumes that words with a strong statistical relationship should be parsed or produced in linear proximity. A comprehension-oriented version of the theory explains information locality in terms of memory and prediction, whereby the greatest prediction benefits accrue to pairs of words with higher mutual information, but this benefit may be decreased by short-term memory deterioration while waiting for intervening words (Futrell & Gibson & et al. 2020; Hahn et al. 2021). An alternative, production-oriented explanation proposes that pairs of words with high mutual information offer an efficiency benefit if they are stored and retrieved holistically, rather than separately, and that such holistically retrieved combinations are preferentially linearised as a contiguous chunk (Mansfield & Kemp in press).

One challenging and intriguing dimension of locality theory is that it clashes, in some instances, with another major theoretical principle, namely 'harmonic' ordering. Harmonic ordering proposes that in each language, head-dependent pairs should tend to be linearised in a consistent direction (Venneman 1974; Travis 1984). This can be explained as a type of grammatical simplicity, avoiding the need for multiple rules specifying word ordering for each specific type of head-dependent pairing. Again, there is psycholinguistic evidence supporting harmony in language production (Culbertson & Franck & et al. 2020; Culbertson & Kirby 2022), and typological evidence suggesting that most languages favour harmonic ordering (Dryer 2018), as already suggested by Greenberg's (1963) earlier proposals. As we will see below, harmony has also found new forms of evidence in phylogenetic modelling of language families.

The clash between locality and harmony occurs whenever there are multiple dependents of a single head (Temperley 2008; Gildea & Temperley 2010). The locality preference is best satisfied if these are arrayed evenly on each side of the head, e.g. $D_1$-$D_2$-H-$D_3$-$D_4$, but the harmony preference is best satisfied if they are all arrayed on the same side, e.g. H-$D_1$-$D_2$-$D_3$-$D_4$. Thus to the extent that heads have multiple dependents, the two principles violate each other. This apparent competition is discussed in one of the dependency locality studies mentioned above (Jing et al. 2022), where it is proposed as one possible explanation for exceptions to locality. This may provide another example of how different languages find different solutions, within the limits of competing principles. This general approach, one that views language as finding optimal solutions under systems of constraints, is a promising quantitative method for modelling linguistic diversity.

## 2.5. Grounding in semantics

In the previous sections we have focused on formal principles of symbolic representations: how are they ordered, stored and produced? But much more could be said about what these symbols *mean*, which is tied to the question of how they reflect our need to distinguish concepts (Kemp et al. 2018), pre-linguistic conceptual structures (Gärdenfors 2000) and embodied perceptual or motor experiences (Feldman 2006).

Information-theoretic models appear again in semantic research. Various studies take well-defined domains such as colour (Zaslavsky et al. 2018), kinship (Kemp & Regier 2012), personal pronouns (Zaslavsky et al. 2021) or tense markers (Mollica et al. 2021), and develop corpus-based metrics for how much information is communicated by the lexemes or grammatical categories in these domains, such as the green-blue colour region, or the temporal region including present and future time. In this approach, every language faces a tradeoff between expressive accuracy, and coding simplicity. Such studies have consistently found that while languages carve up these domains in different ways, the diverse attested partitions show a strong tendency to divide the space in a way that is efficient with respect to communicative needs.

Much recent work in semantics has focused on iconicity and systematicity in language, i.e., the degree to which the form of words, or the relations between words, reflect our perceptual representations of concepts (Dingemanse et al. 2015; McLean et al. 2023). This is one sense in which symbols are 'grounded' in various non-linguistic dimensions of cognition, or in other words, our more basic experiences of the world. Iconicity is another area where more rigorous, and in some case wide-ranging, studies have become possible. For example, one recent study analyses thousands of languages to show that phonological form correlates with concepts in some domains, such as the /n/ phoneme for the concept NOSE, and high vowels for small things (Blasi et al. 2016). However, in follow-up work that has investigated these issues from an information-theory view, there was found to be relatively little mutual information between the average word and its meaning, despite some systematic correspondences (Pimentel et al. 2021).

Representations may also be grounded in non-linguistic perceptions of animacy and causality. These fundamental notions have long been noted as recurrent structuring themes in several aspects of natural language, such as morphological marking (Witzlack-Makarevich et al. 2016), grammatical roles (Dowty 1991) and word-order patterns. Agent precedence is the dominant theme here, with recent experiments showing that comprehenders expect agents to be mentioned first (Bickel et al. 2015), even in a language like Aiwoo, where there are grammatical constraints putting patients first in many contexts (Sauppe et al. 2023). This suggests an important role for agency in our linguistic cognition; but there is also evidence that this is based on pre-linguistic notions. Pre-linguistic infants already pay great attention to animate agents (Adibpour & Hochmann 2023), and non-human primates show the same tendency (Brocard et al. 2024). Some neuro-linguistic work suggests a neural overlap between the representation syntax and agentive actions (Kemmerer 2012). Together these studies suggest that agent precedence, a core principle of contemporary human language, may be based on a neuro-cognitive functional pattern with origins on an evolutionary timescale.

## 3. LINGUISTIC COMPUTATION IN SUB-SYMBOLIC FRAMEWORKS

In the previous sections we have framed linguistic theory as being primarily about the symbolic representations that characterise human language, focusing on principles of representational structure such as efficiency, locality and conceptual grounding. But another important theoretical question, of particular salience in recent years, is how these symbolic structures might be based on 'sub-symbolic' representations. If you open up someone's head, rather than symbols, you will find neuronal assemblies. If you open up contemporary language technology you will not find symbols, but rather numerical calculations wired together in artificial neural networks. So what is the relationship between our symbolic representations, and these sub-symbolic networks?

The relationship between symbolic and sub-symbolic structures first rose to prominence in the 1980s, with the rise of the connectionist modeling paradigm (Rumelhart & McClelland 1986; Elman et al. 1996). We have already briefly mentioned connectionism above, for its role in implementing and evaluating models of grammar. Instead of computing symbolic inputs directly into symbolic outputs (as in, for example, a formal phrase structure grammar), connectionist models mediate computation with artificial neural network structures, that is to say, multiple layers of in/out functions fed on numbers. These intermediate, or 'hidden' layers allow for very powerful statistical learning behaviour, though the nature of the intermediate representations is not always itself interpretable. Within the connectionist literature, the question of symbols might be represented in a distributed fashion across sub-symbolic nodes was soon recognised as a central question (Holyoak & Hummel 2000).

In recent years, connectionist modelling has achieved renewed interest for at least two reasons. First, increasing amounts of data from the brain, and increasing ability to process and run computations on this data, has made it possible to find new links between linguistic theory and the sub-symbolic neural substrates (Schrimpf et al. 2021; Hale et al. 2018), see also [Neurolinguisitcs chapter, this volume]. Second, large language models —the descendants of the original connectionist modeling framework— have achieved impressive performance on many language-related tasks. They are able to produce fluent, humanlike language outputs, without relying overtly on any of the symbolic representations that form the bedrock of linguistic theory.

### 3.1. Vector semantics

Before the rise of modern natural language processing technologies, there was significant work developing non-symbolic statistical representations of word meaning (Jurafsky & Martin 2023). These approaches represented words as vectors of numbers, giving rise to the term *vector semantics.* They were grounded in earlier observations, dating back to the 1950s, that the meaning of a word is related to the contexts in which it occurs (Harris 1954; Firth 1957). By keeping counts of common word-context pairs, it was found that many symbolic-like features could be learned. For example, the word2vec system (Mikolov et al. 2013), induced representations that were similar to gender features. However, vector semantic approaches struggled to learn many of the most interesting features that could be described using symbolic systems, such as quantification, scope, selectional preferences, and, most importantly, compositionality.

## 3.2. Implications of large language models

Compared to vector semantics, large language models (LLMs) represent a more advanced level of language technology based on sub-symbolic structures. LLMs have gained massive public attention due to their use in commercial AI products, but they may also have implications for linguistic theory. Although the sheer size, and some other features, are new, LLMs are again (like our theoretical principles above) the continuation of a longer tradition, namely connectionist modeling.

LLMs are algorithms trained to predict the next word, given a previous history of words, using similar principles to other neural networks. The model begins from a state where all network nodes are assigned random weights, producing error-ridden outputs. But these outputs are compared against target data, with errors used to incrementally adjust node weights, via a process of 'back-propagation' from the output layer to the hidden layers. One advantage of the neural network architectures that form the basis of LLMs is that one can dramatically improve their capabilities through scaling, i.e., increasing the size of the model, or the amount of data on which it is trained. Because of this, LLMs are trained on large quantities of text—far more than a human will ever see in their entire lifetimes—although more data-efficient models are one current area of active research (Warstadt et al. 2023; Hosseini et al. 2024). One major concern with LLMs has to do with data quality. The internet scrapes that form the basis of most LLM's training data include large amounts of low-quality, poorly formed, and biased text. Training models on higher-quality data has been shown to improve model efficiency and produce better outputs (Gunasekar et al. 2023).

The rise of LLMs has given rise to two, interconnected, sets of questions concerning their implications for linguistics and linguistic theory. The first question asks to what extent LLMs learn the symbolic representations (or approximations thereof) posited by linguists, in order to obtain fluency on language-related tasks? These sorts of questions are exciting from the perspective of theory evaluation. If it can be shown that a flexible learning algorithm, capable of approximating any function, learns to represent something similar to any particular linguistic theory, then this would provide compelling evidence that such an analysis is possible and even computationally efficient, given the learning data. For example, Hewitt and Manning (2019) find that models learn approximations of dependency trees over the course of their training. One of the key developments in LLMs is the use of 'attention heads', a type of network mechanism that helps more accurately produce a word that depends on some previous word, such as the object *plans* depending on the verb *discuss* in the sentence, *It declined to <u>discuss</u> its <u>plans</u>*. In subsequent work, Manning et al (2020) show that LLMs develop attention heads that approximately represent particular types of dependency, such as verb-object dependencies. This suggests that the sub-symbolic internal structure of weighted nodes learns a kind of syntactic structure, similar to one proposed by linguists. Another study finds that models develop specific units for tracking things like grammatical number features, and hierarchical phrase structures (Lakretz et al. 2019).

In addition to probing the internal states of models, a related line of work inspects the outputs of models, and asks whether their behavior is consistent with human behavior, in particular with learning abstract language-like generalizations, across a range of test suites. Of these, studies have looked at how consistent models are with human syntax judgments (Marvin & Linzen 2018; Gauthier et al. 2020; Warstadt et al. 2020), pragmatics (Jeretic et al. 2020), and various semantic and commonsense phenomena (Wang et al. 2019; Ivanova et al. 2024). For example, one such test suite might asses models' knowledge of grammar by contrasting the two following sentences:

(2)     a. *The author next to the senators saw **herself** in the mirror*.

  b. *The author next to the senators saw **themselves** in the mirror*.

If a model has learned that reflexive pronouns must match with their referent in terms of number features, then it should find (2a) more probable than (2b). These types of studies have found that models are largely in line with human behavior, and that, especially when it comes to syntax, some of today's state-of-the-art models are within a hair's breadth of being able to emulate people on grammaticality judgment tasks.

  While LLMs clearly approximate human language in several striking ways, there is great disagreement about the implications for linguistic theory. Below, we provide three possible responses to this question. Judging between these responses is an active, and important area of linguistic theory research. The first response is largely critical and assumes that language models are unlikely to meaningfully contribute to linguistic theory. Reasons to be skeptical of LLMs' impact include classic arguments about the difference between probability distributions produced by models and grammaticality (Chomsky 1957; Clark & Lappin 2010). Other arguments point out that LLMs don't have intention, which is necessary for natural language understanding (Bender & Koller 2020), and that LLMs will learn any language, including the impossible ones (Moro et al. 2023; but see also Kallini et al. 2024). In the next section, we will come back to the theme of grounding in intentional agents. Separate from the theoretical critique of LLMs, there is also the important issue of their environmental impact, as their training process uses massive amounts of electricity (Strubell et al. 2019; Patterson et al. 2021).

  On the opposite end of the spectrum is the perspective that LLMs represent the possibility of a theoretical paradigm shift (Baroni 2022; Piantadosi 2024). The idea here is that LLMs do a much better job of predicting and emulating human linguistic behavior than symbolic theories, and therefore should be taken seriously as theoretical proposals themselves. Just as our theory of hurricanes includes a statistical model that can accurately describe weather systems' behavior, a successful theory of language could be similar.

  Somewhat in the middle is the perspective that LLMs, while not instantiating theories themselves, represent a major step forward in our ability to test and refine existing symbolic and information-theoretic proposals. Already in this chapter we have discussed several examples where information-theoretic proposals are tested against naturally occurring language data. Several of these experiments relied on LLMs to provide accurate estimates of in-context probability that would not have been possible, at least to the desired level of accuracy, without these new tools. One area where LLMs stand to have significant impact which we have not discussed is in language acquisition. Here, researchers have long investigated the interplay between childrens' innate learning biases and the primary linguistic data from which they learn (Clark & Lappin 2010; Becker & Ud Deen 2020). Although the gold standard to investigate such questions is a controlled experiment that modulates the input available to a child, such studies would be unethical to conduct. However, this is not true for LLMs, and there is already an active field of research using LLMs to home in on what elements of language data, and inductive learning biases are necessary for in-principle learning of human language (Dupoux 2018; Warstadt & Bowman 2022).

  It is the third, middle path, for which we advocate. As we have already proposed in previous sections, advances in language data and computational methods can benefit linguistic theory by providing more explicit and accurate evaluation of theoretical proposals. LLMs have already been useful for testing and refining various theories, and have contributed towards a better integration of previous-generation linguistic theories with theoretical frameworks from allied disciplines, most notably information theory.

Irrespective of the position one takes on whether LLMs mechanisms are comparable to human cognition, or completely incomparable, they unquestionably add to our methodological toolkit for testing theories.

## 4. INTERACTING MINDS

As was mentioned above, one way in which LLMs differ radically from human language is that their computational processes are not grounded in intentional agents. LLM language production is, to a large extent, simply a (very good) word prediction algorithm, but human language production is instead embedded in a field of perceptions, beliefs and intentions that go beyond language. Furthermore, the human subject also has rich representations of their interlocutors' perceptions, beliefs and intentions. Actual language is not just subjective, but *intersubjective*.

We began this chapter by presenting linguistics as a theory of symbolic representations. But language cannot be equated with disembodied symbols on the page or screen. That's just annotation. Language occurs as actions between interlocutors, who respond to each others' outputs and do enormous amounts of inferential work (Grice 1975; Levinson 2000). The inferences involve modelling the subjective state of the interlocutor (*what are they thinking*), and modelling this in a recursive manner (*what they are thinking that I am thinking that they are thinking…*) While studying dismebodied representations may be a useful abstraction, it is therefore far from a complete theory of language. And it is increasingly clear that grammatical structures, from word order to inflection to complex clause types, are to a large extent designed not just to represent the state of the world, but also to negotiate our intersubjective representations.

Pragmatics is a late-comer to linguistic theory. While phonology, morphology and syntax follow centuries or millennia of analysis (Seuren 1998), pragmatics only passed more recently from philosophy (e.g. Wittgenstein 1953; Austin 1962) into empirical linguistic research. And its importance is arguably still unfolding, for example in the degree of attention it receives in descriptive grammars and fieldwork. The increased awareness of intersubjectivity has facilitated 'prag-pilled' grammatical descriptions that recognise, for example, inflectional categories that primarily mark states of speaker and addressee knowledge (Evans et al. 2018), rather than more traditional core categories such as tense or modality. The increasing documentation of languages with evidential and intersubjective verb inflections has also highlighted the importance of intersubjectivity (Bergqvist & Grzech 2023). Similarly, there is increasing work on languages for which clausal constituent order is driven primarily by intersubjective inferences, rather than semantic roles such as Agent and Patient (Mithun 1992; Nordlinger et al. 2022; Ma & Mansfield 2024). But again, this function is still marginalised by approaches that model clausal constituent order in terms of semantic roles.

One issue for researching intersubjectivity is the nature of the data. While all areas of linguistic research once relied heavily on introspection, other areas have long since moved on to more sophisticated methods. Pragmatics has arguably been even more reliant on introspective or impressionistic judgements about the intersubjective states associated with utterances. One approach to improving this has been conversation analysis, which involves careful qualitative analysis of natural conversation data, encompassing not just spoken words but also timing, voice quality, gesture, gaze etc. (Floyd 2021). Another approach has been the development of focused elicitation tasks that prompt speakers to make rich use of intersubjective marking in contexts that are revealing to the researcher, such as contrasts between shared and exclusive information (Knuchel 2020). But perhaps the most important development is in experimental paradigms that allow explicit hypothesis testing, directly comparing responses in such a way that reveals intersubjective representations (Schwarz 2017).

One major topic in the study of intersubjectivity, which has been targeted for experimental research, is the nature of reference. How do speaker and addressee align joint attention on the same referent? Of particular interest are personal pronouns, notable for their frequent use and deictic semantics. A series of experiments show that pronoun use and interpretation depends on multi-faceted and subtle principles of inference and intersubjective representation (Kehler & Rohde 2013; Kehler & Rohde 2019). One of the most compelling aspects of this research is that it develops a mathematical model of speakers' and addressees' probabilistic expectations, that is to say, how they respond to language use by adjusting their estimates of their interlocutors' mental representations.

Mathematical modelling of intersubjectivity has been further developed in the Rational Speech Act framework (RSA: Frank & Goodman 2012), in which the study of intersubjectivity is united with the information-theoretic trend identified above. In the RSA model, listeners interpret utterances by reasoning about the utility that the utterance would have to a speaker. They assume that such a speaker, in turn, is trying to efficiently communicate a message to a listener, and attempts to select an utterance from a range of possible options. The idea is that each layer of the model is reasoning about how to communicate efficiently, with respect to the next layer down. Language producers and comprehenders are 'rational' in the sense that they aim for communicative utility, and they assume that their interlocutors share this rational approach. By stacking multiple RSA layers on top of each other models can become quite complex and demonstrate interesting emergent behavior. Simulations have demonstrated that these basic building blocks (efficient communication; recursive reasoning) can give rise to a wide array of pragmatic phenomena (Degen 2023). That being said, not all types of pragmatic phenomena can be elegantly captured by RSA models (Cremers et al. 2023), and RSA models tend to have many free parameters, or parameters that are difficult to interpret from a psychological perspective. One major shortcoming of RSA models is that they are developed to explain toy domains, and operate over a limited set of meanings and alternative utterance options. Expanding RSA models to cover more naturalistic data is a challenging but necessary next step. The promise of such developments would be a more explicit model of human language use, taking account of the intersubjective inferences that are typically excluded from disembodied theoretical representations.

## 5. EVOLUTIONARY LINGUISTICS

Intersubjectivity is just one type of embodiment that linguists have often strategically ignored to make useful abstractions. Another way in which we usually deal in abstractions is in the idealisation of a homogeneous competent speaker (Chomsky 1965), setting aside the dynamics of language change, and the variation of individuals in a speech community. Of course, the study of language change has a long tradition in historical linguistics, while the study of individual variation has been of central interest in sociolinguistics (Weinreich et al. 1968). However these research traditions have remained largely separate from mainstream theoretical work on linguistic representations. Chomsky's distinction between competence and performance, following Saussure's earlier distinction between *langue* and *parole*, were successful in circumscribing the formal study of symbolic structure. But many linguists since have aimed to bridge the divide between discrete symbols and historically contextualised, socially motivated agents.

The recent rise of 'evolutionary' linguistics can be seen as an attempt to bridge the gap. Like historical linguistics or socio-linguistics, it is concerned with the dynamics of variation and change, but it differs from both in taking a wider spatial and temporal scope: how have the dynamics of variation and change played out across the globe, over thousands of years? How did human languages evolve from earlier systems of animal communication? With its concern for cross-linguistic comparison and coverage, evolutionary linguistics can also be seen as a development of typological linguistics, but with an increased focus on dynamic process and deep history (Trudgill 2011; Di Garbo et al. 2021). Adding to this the dimension of cross-species comparison, evolutionary linguistics aims to integrate biological sciences with linguistics, cognitive science and neuroscience. It is an ambitious project, but the pace of recent developments is impressive.

If linguistic theory posits symbolic representations, evolutionary linguistics challenges us to consider how such representations may have developed from hominid antecedents, transmitted over thousands of generations up to the present. While biological transmission via DNA has held primacy as the canonical object of evolutionary research, anything that is replicated with modification may be analysed in an evolutionary lens, including a wide range of culturally transmitted behaviours (Hull 1980; Dawkins 1982; Croft 2000). Arguably, language is the most complex example of a culturally transmitted behaviour, though this is not to deny the importance of biological adaptations, captured in a 'co-evolutionary' perspectives that sees culture and biology shaping one another (Durham 1991).

But there is also a paradox in the evolutionary perspective on language. 'Languages' do not evolve and reproduce themselves, as if they were organisms. Instead, language users (speakers and signers) reproduce linguistic structures, which they learn and transmit over many generations. Linguistic utterances are approximate replications of other utterances, and these can be seen as the proper analogue of a 'population' composed of variants and transmissible features (Croft 2000), though these populations of utterances do not form parent-offspring lineages in the manner of DNA (Godfrey-Smith 2009: 162). This implies that 'languages' are not discrete entities, and are not inherited as a monolithic package from one generation to the next or from one individual to the next. In this spirit, some recent work in evolutionary linguistics explores the ways in which different elements of language are transmitted. For example, one can distinguish rates of change in grammar versus basic lexicon, yielding the perhaps surprising result that grammar appears to change faster than vocabulary, in Austronesian and Indo-European phylogenies (Greenhill et al. 2017). Another study breaks languages of north-east Asia down into their vocabulary, grammar and phonological components, to test which of these correlate most strongly with long-term population histories as evidenced from genetics (Matsumae et al. 2021). Here it was found that grammar correlates with population history more closely than either vocabulary or phonology. This may appear to contradict the previous study, by indicating a slow rate of change for grammar. But in fact the studies are not directly comparable, as the former traces change over language family trees (which are themselves based on sound-changes and vocabulary), while the latter is based on population history, which sometimes diverges from language family history (Barbieri et al. 2022). In short, much more research is needed in this complex area.

Despite the demonstrable decoupling of vocabulary, phonology and grammar, and the actual variation of individual usage, 'languages' nonetheless seem to mostly cohere (Enfield 2014). Put another way, although we know that speech communities are porous to influence of neighbouring communities, the tree model of language inheritance still seems to work, even though it treats 'languages' as entities that each inherit via a single path from earlier states. As we will see in the following section, phylogenetic research based on the traditional tree model is currently an area of vibrant research activity, though there are also methodological investigations of how we can move away from a strict tree model.

## 5.1. Advances in phylogenetic methods

Major methodological advances in evolutionary linguistics focus on the use of computational phylogenetic models (Gray et al. 2013; Carling et al. 2022). These are statistical models of branching tree structures, typically involving languages as leaf nodes, with branches connecting these languages to imputed or attested ancestor languages. The tree consists of a 'topology', that is the branching structure connecting languages, a set of linguistic features that are assigned values for every language node in the tree, and a process of change that takes place along the branches. Thus this reflects the fundamental 'tree model' developed in historical linguistics and evolutionary biology, allowing it to be applied as a formal mathematical model. One particularly attractive feature of such modelling is that it permits explicit modelling of uncertainty, which is often a major feature of historical reconstruction (Goldstein 2022).

Phylogenetic modelling is complemented by ever-growing public databases. Tree structures are now standardised and accessible in Phlorest (Forkel & Greenhill 2023); lexical data for thousands of languages is accessible in sources such as Lexibank (List et al. 2022) and the Automated Similarity Judgement Program (Wichmann et al. 2020); phoneme inventories in Phoible (Moran & McCloy 2019), and grammatical features in Grambrank (Skirgård et al. 2023), the World Atlas of Language Structures (Haspelmath et al. 2005), and AutoTyp (Bickel et al. 2022). These and other resources are promoting more open and reproducible methods in evolutionary linguistics.

The most linguistic feature that is most often modelled phylogenetically is the lexicon, though phonological and grammatical features can also be modelled (Dunn et al. 2008). For example, one can treat the presence/absence of cognates as a changing (discrete binary) feature, then infer a model of how cognates are gained and lost through lexical replacement in a language family. These phylogenetic models can be constrained by a certain topology, typically established through the traditional comparative analysis of sound changes, or by dates where available from historical and archaeological sources. In this case a model might infer rates of change, impute likely features to unobserved ancestor languages (Carling & Cathcart 2021) and estimate dates of nodes for which other historical information is not available (Posth et al. 2018; Heggarty et al. 2023; Greenhill et al. 2023). But where less is known about a language family, statistical inference can also be used to calculate the likelihood of various topologies as well as the likely chronology (Bouckaert et al. 2018; Kolipakam et al. 2018). Phylogenetic modeling is thus considerably expanding our insights into the deep history of languages and their evolutionary dynamics. The inherent uncertainty in such models, rather than being a weakness, can be seen as a more scientific approach compared to traditional argumentation where each scholar stakes their reputation on a particular inferred scenario.

While phylogenetic models are by definition based on the branching tree structure, there are also recent explorations of how the the non-treelike dimensions of language structure can be modelled (cf. Meakins 2022). Some studies attempt to integrate a vertical phylogeny with various types of 'horizontal' process, which represent independent forms of transmission, such as contact between attested languages, or residual influence from unattested and otherwise unknown languages that may have existed in the past (Kalyan & François 2018; Neureiter et al. 2022; Guzmán Naranjo & Mertner 2023; Efrat-Kowalsky et al. 2024). This is a relatively new area of research, but already suggests that future computational work on language history will move beyond the strict branching tree.

The dynamics of change inferred from phylogenetic models reflect in interesting ways on the principles of linguistic structure discussed above. If we propose that certain types of structure are favoured by linguistic computation and communication, then we might find reflections of this in evolutionary dynamics (Maslova 2000). Such reflections have indeed been found in some recent studies.

For example, a recent study infers phylogenies for a large sample of 1,626 languages with word order data, focusing on the estimated probabilities with which one word order changes to another in these phylogenies, and the correlations between such word orders (Jäger & Wahle 2021). This method reveals certain word-order correlations as universal statistical preferences across phylogenies, many of which are in striking conformity with long-proposed typological correlations (Greenberg 1963; Dryer 1992). In terms of the structural principles mentioned above, these universal correlations are compatible with the notion that harmonic dependencies are a preferred method of symbol linearisation. Another example of phylogenetic inference relating to structural principles is a study of the agency/animacy preference. As noted above, this implies that ergative grammar, which marks agents as unexpected referents, should predominantly occur in unusual situations, for example with inanimate agents. Phylogenetic inference about case systems in the Pama-Nyungan family, where ergativity is unusually common, indeed suggest that the most robust type of ergativity involves exactly these types of restricted, animacy-conditioned systems (Phillips & Bowern 2022).

## 5.2. Evolution among individuals

While phylogeny models each language as a holistic system, there is also current work modelling language as the aggregate of multiple individuals, who vary in their idiolects. The work of Simon Kirby and colleagues (e.g. Kirby 2001; Kirby et al. 2008) has been particularly influential here, exploring how individuals may converge on usable shared codes. Earlier work involved purely computational simulations, aiming to create an explicit model of how individuals might develop communicative norms. In general, work of this type is known as 'agent-based' modelling, though this generally implies only that there is some representation of individuals, rather than implying the type of intentional, subjective representations discussed in the previous section.

However further agent-based research has involved actual humans. While studies with LLMs attempt to make artificial systems perform human-like actions, some agent-based evolutionary research makes humans perform artificial things, in particular by testing how they respond to artificial languages. While this inevitably takes us a long way from real language use, it does allow researchers to explore the possibilities of counter-factual symbolic systems, with theoretically interesting properties. In one famous study, individual participants teach each other invented names for simple coloured shapes, starting from an experimental setup where all names are unstructured and random (Kirby et al. 2008). Interestingly, the iterative learning among individuals quickly gives rise to rudimentary forms of linguistic structure. This work connects evolution back to the questions of symbolic structure raised above, but rather than asking 'what does an efficient system look like', in a static sense, it asks, 'how can efficient (or learnable) system arise?' (Kirby et al. 2015; Tamariz & Kirby 2016).

Other agent-based models have explored natural language data, for example investigating historical changes in grammatical structure, and arguing that these can be better explained by individuals who vary in their production during their lifetimes, as opposed to a simpler model where each generation either replicates or modifies the pattern used by the previous generation (Blythe & Croft 2021; see also Baxter & Croft 2016) . Other work has applied the Fisher-Wright model, a mainstay of genetic evolutionary theory, using it to test for biases in inter-generational transmission among individuals (Meakins et al. 2019; Hua et al. 2021). While this type of individual-based modelling for diachronic data remains preliminary, we expect this to be an area of further growth in evolutionary linguistics.

# 6. THE FUTURE OF LINGUISTICS

We can only hope that the coming decades will be kind to this chapter. If we are fortunate enough to have readers thirty years from now, no doubt some of our points will seem quaint or misguided. However we believe that most of the trends identified above will either become norms of linguistic research, or will be identifiable as foundations upon which other approaches have built.

Early in the chapter we mentioned the balance between theory and methods. Linguists should use rigorous methods drawing on broad, high-quality datasets, but this will never replace the need for theoretical insight, giving us the ability to identify interesting questions, formulate plausible hypotheses, and translate these into explicit representations and measures. The current drive towards ever-greater data, and sophisticated computational or quantitative methods that are at times hard to follow, does pose a risk to linguistics. There is a risk that high-profile, big-data studies contain assumptions and confounds that most linguists would reject as theoretically implausible. But we expect that in the coming decades, linguists will become ever more familiar with computational and quantitative techniques, and this will gradually bring about more consensus and transparency on appropriate methods.

Finally, while we are confident that scientific study of language has a bright future, it is unclear how linguistics in thirty years will be related to the linguistics of the present day. Linguistics is a relatively modern academic unit, having been born from a confluence of philology, anthropology and computer science in the mid twentieth century. We cannot assume that this confluence will persist in the same form. In this chapter we have emphasised the 'language science' face of linguistics, but there are also many linguists who see themselves primarily as humanities scholars, critical theorists or education researchers. An important future challenge is to maintain connections between these approaches, while also fitting into university structures that tend to separate science and humanities.

Mollica, Francis & Bacon, Geoff & Zaslavsky, Noga & Xu, Yang & Regier, Terry & Kemp, Charles. 2021. The forms and meanings of grammatical markers support efficient communication. *Proceedings of the National Academy of Sciences*. Proceedings of the National Academy of Sciences 118(49). e2025993118. (doi:10.1073/pnas.2025993118)

Moran, Steven & McCloy, Daniel (eds.). 2019. *PHOIBLE 2.0*. Jena: Max Planck Institute for the Science of Human History. (https://phoible.org/)

Moro, Andrea & Greco, Matteo & Cappa, Stefano F. 2023. Large languages, impossible languages and human brains. *Cortex* 167. 82–85. (doi:10.1016/j.cortex.2023.07.003)

Neureiter, Nico & Ranacher, Peter & Efrat-Kowalsky, Nour & Kaiping, Gereon A. & Weibel, Robert & Widmer, Paul & Bouckaert, Remco R. 2022. Detecting contact in language trees: a Bayesian phylogenetic model with horizontal transfer. *Humanities and Social Sciences Communications*. Palgrave 9(1). 1–14. (doi:10.1057/s41599-022-01211-7)

Nordlinger, Rachel & Rodriguez, Gabriela Garrido & Kidd, Evan. 2022. Sentence planning and production in Murrinhpatha, an Australian "free word order" language. *Language*. Linguistic Society of America 98(2). 187–220.

Patterson, David & Gonzalez, Joseph & Le, Quoc & Liang, Chen & Munguia, Lluis-Miquel & Rothchild, Daniel & So, David & Texier, Maud & Dean, Jeff. 2021. Carbon Emissions and Large Neural Network Training. arXiv. (doi:10.48550/arXiv.2104.10350) (http://arxiv.org/abs/2104.10350)

Phillips, Joshua & Bowern, Claire. 2022. Bayesian methods for ancestral state reconstruction in morphosyntax: Exploring the history of argument marking strategies in a large language family. *Journal of Language Evolution* 7(1). 1–15. (doi:10.1093/jole/lzac002)

Piantadosi, Steven T. 2024. Modern language models refute Chomsky's approach to language. (https://lingbuzz.net/lingbuzz/007180)

Piantadosi, Steven T. & Tily, Harry & Gibson, Edward. 2011. Word lengths are optimized for efficient communication. *Proceedings of the National Academy of Sciences* 108(9). 3526–3529. (doi:10.1073/pnas.1012551108)

Pimentel, Tiago & Meister, Clara & Wilcox, Ethan & Mahowald, Kyle & Cotterell, Ryan. 2023. Revisiting the optimality of word lengths. *Proceedings of the 2023 Conference on Empirical Methods in Natural Language Processing*, 2240–2255. (doi:10.18653/v1/2023.emnlp-main.137)

Pimentel, Tiago & Roark, Brian & Wichmann, Søren & Cotterell, Ryan & Blasi, Damián. 2021. Finding Concept-specific Biases in Form–Meaning Associations. In Toutanova, Kristina & Rumshisky, Anna & Zettlemoyer, Luke & Hakkani-Tur, Dilek & Beltagy, Iz & Bethard, Steven & Cotterell, Ryan & Chakraborty, Tanmoy & Zhou, Yichao (eds.), *Proceedings of the 2021 Conference of the North American Chapter of the Association for Computational Linguistics: Human Language Technologies*, 4416–4425. Online: Association for Computational Linguistics. (doi:10.18653/v1/2021.naacl-main.349)

Posth, Cosimo & Nägele, Kathrin & Colleran, Heidi & Valentin, Frédérique & Bedford, Stuart & Kami, Kaitip W. & Shing, Richard et al. 2018. Language continuity despite population replacement in Remote Oceania. *Nature Ecology & Evolution*. Nature Publishing Group 2(4). 731–740. (doi:10.1038/s41559-018-0498-2)

Rizzi, Luigi. 2013. Locality. *Lingua* (SI: Syntax and Cognition: Core Ideas and Results in Syntax) 130. 169–186. (doi:10.1016/j.lingua.2012.12.002)

Rumelhart, David E. & McClelland, James L. 1986. *Parallel distributed processing, volume 1: Explorations in the microstructure of cognition*. Cambridge, MA: MIT Press. (https://mitpress.mit.edu/books/parallel-distributed-processing) (Accessed March 30, 2018.)

Sauppe, Sebastian & Naess, Åshild & Roversi, Giovanni & Meyer, Martin & Bornkessel-Schlesewsky, Ina & Bickel, Balthasar. 2023. An agent-first preference in a patient-first language during sentence comprehension. *Cognitive Science* 47(9). e13340. (doi:10.1111/cogs.13340)

Schrimpf, Martin & Blank, Idan Asher & Tuckute, Greta & Kauf, Carina & Hosseini, Eghbal A. & Kanwisher, Nancy & Tenenbaum, Joshua B. & Fedorenko, Evelina. 2021. The neural architecture of language: Integrative modeling converges on predictive processing. *Proceedings of the National Academy of Sciences*. Proceedings of the National Academy of Sciences 118(45). e2105646118. (doi:10.1073/pnas.2105646118)

Schwarz, Florian. 2017. Experimental pragmatics. *Oxford Research Encyclopedia of Linguistics*. (doi:10.1093/acrefore/9780199384655.013.209)

Seuren, Pieter A.M. 1998. *Western linguistics: An historical introduction*. Oxford: Blackwell.

Shannon, Claude E. 1948. A mathematical theory of communication. *Bell System Technical Journal* 27(3). 379–423.

Skirgård, Hedvig & Haynie, Hannah J. & Blasi, Damián E. & Hammarström, Harald & Collins, Jeremy & Latarche, Jay J. & Lesage, Jakob et al. 2023. Grambank reveals the importance of genealogical constraints on linguistic
23